\documentclass[letterpaper, 10 pt, conference]{ieeeconf} 

\IEEEoverridecommandlockouts                              
                                                          
\usepackage{cite}
\usepackage{algorithmic}
\usepackage{graphicx}
\usepackage{textcomp}
\usepackage{xcolor}
\usepackage{booktabs}
\usepackage{microtype}
\usepackage{array}
\usepackage{booktabs}
\usepackage{textcomp}
\usepackage{float}
\usepackage[hyphens]{url}
\usepackage{hyperref}
\usepackage{breakurl}
\usepackage{subcaption}

\usepackage[font=small,skip=2pt]{caption}
\def\BibTeX{{\rm B\kern-.05em{\sc i\kern-.025em b}\kern-.08em
    T\kern-.1667em\lower.7ex\hbox{E}\kern-.125emX}}
    
\begin{document}
\setlength{\textfloatsep}{6pt}
\setlength{\dbltextfloatsep}{6pt}
\setlength{\floatsep}{6pt}

\title{\LARGE \bf{Bridging Teacher Expectations and Robot Learning via Coupling Dynamics}
\author{Evan Dallas$^{1}$, Sean Dallas$^{1}$, and Wing-Yue Geoffrey Louie$^{1}$}

\thanks{$^{1}$E. Dallas,  S. Dallas, and W. Louie, are with the Department of Electrical and Computer Engineering, Oakland University, Rochester, MI 48309, United States {\tt\small{edallas, srdallas, louie}@oakland.edu}}
\thanks{This work was supported by the National Science Foundation CAREER under Grant \#2238088}}

\maketitle

\begin{abstract}
Human-robot teaching focuses on enabling non-technical experts to customize robots according to their needs after deployment. With recent advances in machine learning, human-robot teaching is no longer confined to offline learning where the data gathering step from a human teacher is separated from when the robot learns. Instead, more recent approaches for human-robot teaching focus on coupling human teaching with robot learning. This coupling impacts the structure, timing, and content of the teaching and learning interaction. However, it is currently unclear how such coupling dynamics affect human-robot teaching effectiveness and human perceptions towards the teaching process. Informed by human learning theories, in this paper we propose a new scale for classifying human-robot teaching interactions according to coupling dynamics present between the human teacher and robot learner. We apply this scale to a subset of the human-robot teaching literature to identify how coupling dynamics and human teacher mental model mismatches with the ground truth robot learning system affect teaching effectiveness and human perceptions towards the teaching process. 
\end{abstract}

\section{Introduction}



Robots have demonstrated significant positive impacts in a range of applications. However, a major barrier for their adoption by the public is the perception that they are “single-use” tools, cannot adapt to the rapidly changing diverse needs of users \cite{sochanski2021therapists, conti2017robots, alcorn2019educators, zubrycki2016understanding}, and concerns over requiring significant expertise or training to be customized. Human-Robot Teaching (HRT) approaches like learning from demonstration are being explored as a way for end-users to customize robots as demonstration is a natural approach for teaching robots \cite{argall2009survey, chernova2022robot}. HRT systems have demonstrated potential in teaching robots to deliver recreation therapy to older adults \cite{louie2020social}, autism therapy to children \cite{tyshka2023interactive, tyshka2022transparent}, and customer service in retail \cite{liu2016data}. 


Historically, a significant barrier in HRT was the mandatory downtime between data collection and assessment. This was due to the prevalence of offline learning (or batch learning), where an agent is trained on a fixed, pre-recorded dataset of expert traces without any real-time teaching interactions. While safe, this approach suffered from small execution errors that would compound, leading the robot into unfamiliar states it had never seen in the static demonstrations, often resulting in catastrophic failure. The transition to online learning was catalyzed by the development of interactive frameworks, most notably Dataset Aggregation (DAgger) in 2011, which allowed robots to proactively query experts for labels during execution \cite{ross2010no}. This shifted the paradigm from passive observation to an iterative dialogue, where the robot learns to recover from its own mistakes by incorporating real-time teacher feedback into its evolving policy.

Despite the improvement from offline learning to online HRT interactions, a critical gap remains within the literature regarding the human element of the HRT interaction. While online HRT frameworks are better than offline implementations for the teaching interaction, they rarely consider that the end user often has no formal training on how the robot actually learns. This means many current systems prioritize the optimization of the robots policy while lacking consideration of the teachers experience level, constraints, and cognitive load. Consequently if the teaching interaction feels unintuitive or taxing, the quality of the demonstrations and by extension the robots learning deteriorates \cite{ayub2025continual}. 

With these recent advancements in HRT models, robot learning is no longer confined to offline or opaque processes, but is increasingly shaped by the structure, timing and content of human teaching \cite{du2018online}. These developments create new opportunities to improve robot learning outcomes through more effective teaching, enhanced pedagogical naturalness, and identifying sources of misalignment between the teacher's and the robot learner's mental model \cite{romat2016natural, nikolaidis2012human, tabrez2020survey}. 

To address this misalignment, we explore established human learning theories that can be adapted to inform the design of robotic learners and their accompanied HRT interactions. Just as these theories are employed by both novice and veteran educators to understand how people acquire, process, and retain information \cite{liu2005vygotsky}, they offer insight into how a robot should solicit and receive knowledge. By drawing from pedagogical frameworks such as scaffolding and constructivism, we investigate how human-robot teaching interactions can be designed to feel more natural to human participants. This alignment has the potential to ensure the robot learning process mirrors the teacher's expectations and better bridges the gap between their respective mental models resulting in a more efficient transfer of knowledge.

Informed by human learning theories, in this paper we propose a new lens for comparing HRT models based on the level of coupling between teaching interactions and learning that characterizes when and how robot learning processes are perceived by the human teacher. This lens is in the form of a 4-point scale that places HRT designs by the frequency of when the teaching interaction and the robots learning update are linked. We then analyze different HRT learning designs using this scale and investigate how the teaching process aligns with a human teachers' mental model of the robot learner. We argue that a misalignment between the level of coupling between HRT interactions and learning contributes to breakdowns in pedagogical naturalness, even in interactions that are otherwise perceived as socially natural. 



\section{Constructivist Learning Theory}

Learning theories are the framework for understanding how knowledge is formed, how teaching should be structured, and how learning is shaped through teaching interactions \cite{illeris2018overview}. Among learning theories, Constructivism is the most widely recognized due to its emphasis on meaning making, interactions, active learning, and context as essential elements to a learning environment \cite{ertmer2013behaviorism,allen2022introduction,jones2002impact}. Krahenbuhl \cite{krahenbuhl2016student} defines Constructivism as an epistemological view in which knowledge is derived in a meaning-making process where learners construct individual interpretations of their experiences to create meaning. A central assumption in Constructivism is that teaching interactions and learning are inseparable. As a result, Constructivism carries expectations on the nature of the timing and visibility of learning as the teaching interaction unfolds. These assumptions become increasingly relevant in HRT, where robot learning systems may not update their knowledge structures in ways that will align with a teacher's expectations \cite{richter2025improving}. The following paragraphs explore two Constructivist principles that are relevant to HRT interactions. 

\subsection{Principle 1}
From a constructivist perspective, experience is directly meaningful to knowledge formation \cite{jones2002impact,hein1991constructivist,caine1991making}. Learning is not assumed to only occur after a teaching interaction has completed, but constructed through engagement with tasks and social partners. This results in educators placing value in each experience, even failed ones. Additionally, educators expect teaching actions such as demonstrations to have a meaningful effect within the teaching interaction. In contrast, robot learning systems treat experience as a data point. This difference generates misalignment because the human teacher understands an experience as learning, whereas the robot interprets the same experience as merely an input.

\subsection{Principle 2} 
Learning in constructivist theories is understood as an incremental process, one that continues to unfold throughout an experience. Prior knowledge structures are continuously refined in response to interaction, rather than updated at discrete boundaries\cite{jones2002impact, hein1991constructivist, caine1991making}. As a result, teachers often expect the learner's behavior to adapt within the course of a teaching interaction through teaching actions such as use of feedback, correction, or repetition to guide this refinement. In contrast, many robot learning systems isolate learning into episodes or phases with delayed or invisible adaptations. This temporal gap creates misalignment where the teacher expects continuous responsiveness, while the robot learning associated with an experience remains temporally segmented. 

The first principle emphasizes the role of experience and timing in knowledge construction and highlights that learning occurs through engagement with tasks. The second principle focuses on the iterative nature of learning, where knowledge structures are continuously refined throughout interaction through reinforcement, correction, and repetition of the learner.


To contrast HRT with human learning theories, we analyze HRT interactions through their pedagogical structure. Pedagogical structure defines the relationship between teacher and learner, and consists of what is taught by the human and how learning is enabled within the HRT interaction \cite{shah2021conceptualizing}. This structure is the basis for how information flows in a HRT interaction and informs the teacher's expectations about how learners should respond when learning is visible \cite{ley2019improving}. 



\section{Human-Robot Teaching Taxonomy} 

\begin{figure}[b!]
    \centering
    \includegraphics[width=0.5\linewidth]{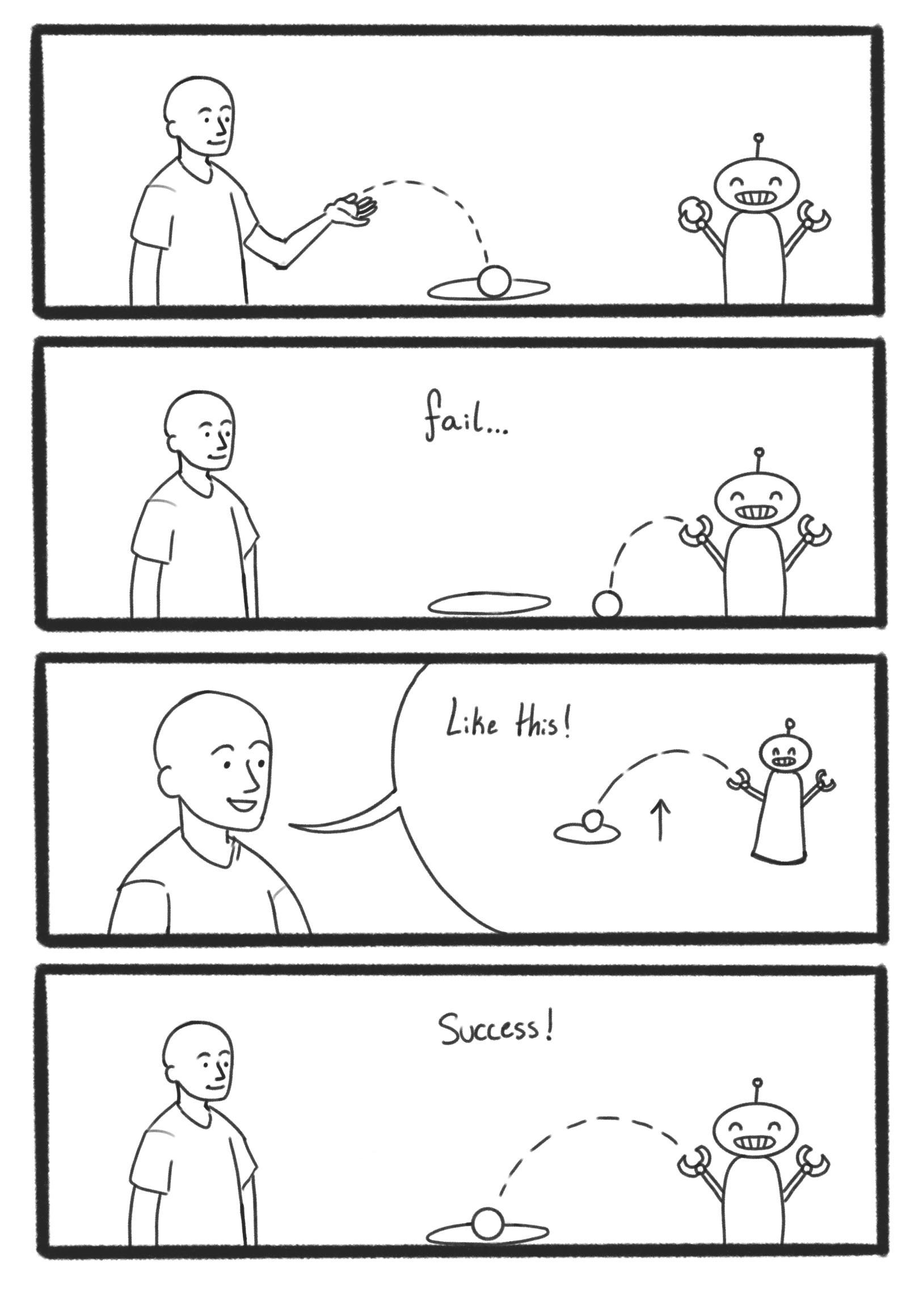}
    \caption{Human-Robot teaching interaction where a human teaches a robot to throw a ball.}
    \label{fig:HRTIFlow}
\end{figure}

Humans perceive teaching as a continuous, meaning-making dialogue, while robotic learning systems often categorize learning through discrete events that usually follows a "perform task" and "then receive feedback" design \cite{ertmer2013behaviorism, schunk1996learning}. Therefore, to bridge the continuous space of human-human teaching to the discrete space of robot learning it's necessary to break down the HRT scenario into components that we can relate between the two paradigms. 

Herein, we define a HRT scenario as the environment to which a HRT interaction takes place, the physical or virtual space where the human teacher and the robot learner converge to exchange knowledge. In this space, the human teacher acts as the source of expertise, responsible for translating their intuitive knowledge into structured guidance via teaching. The robot learner functions as the learner responsible for processing the teaching into actionable learning updates.

The mechanism that enables this exchange is the HRT interaction. This interaction is defined by three pillars, the target competency, the operational task, and the exchange of information. Target competency represents the why of the interaction, the overarching goal or state of success the robot must achieve. To reach this goal, the robot must perform the operational task, which is the specific sequence of actions required by the scenario. Lastly, the exchange of information is the pedagogical bridge used as a corrective signal to close the gap between the robot's current output and the teacher's expectation. In this context, the exchange of information consists of teacher interventions such as teacher feedback, which includes reinforcement, direct corrections, and guided repetition. These interventions serve as a corrective signal, allowing the teacher to iteratively recalibrate the robot’s behavior until its execution aligns with the target competency.

\section{HRT Coupling Scale}

Beyond the physical and roles-based components of the HRT scenario, the nature of the knowledge transfer taking place is defined by the temporal and behavioral dynamics of the learning process. We define the frequency of learning as a construct, which categorizes the density of coupled robot knowledge acquisitions and learning updates. At the baseline, a scenario may involve no learning or a singular instance of learning. As the interaction becomes more complex, it may shift to an iterative model involving multiple discrete updates, eventually reaching step learning. In this state, the robot’s internal learning updates occur so rapidly and consistently that it mimics the fluid, constant progression observed in human-human teaching and learning, appearing to learn at every moment of the interaction. Below we define each level within our scale:

\subsection{No Coupling}

At the first position, no coupling, the robot learner may participate in a teaching interaction which is not used to update its learning system. In this structure, interaction and learning are intentionally decoupled, meaning the robot's knowledge remains unchanged during and after the teaching interaction. This design stands in contrast to how learning is typically expected to occur during human teaching interactions. For example, in a baseline performance trial, a robot may collaborate with a human to move heavy pallets in a warehouse. Even if the human teammate provides verbal feedback or manual corrections to the robot’s positioning, the robot's navigation logic remains static to ensure the experimental data reflects its pre-programmed state rather than an evolving policy.

\subsection{Singular Coupling}

The second position, singular coupling, refers to a set of experiences or a completed interaction followed by a single learning update. The robot's internal knowledge is only updated once, typically during an offline phase after the session has concluded. This is common in batch-learning scenarios where a robot observes a full suite of demonstrations and generates a new behavioral model only after the teacher has finished the entire sequence. 

\subsection{Iterative Coupling}

The third position, iterative coupling, describes a higher frequency of learning within the HRT interaction. Unlike lower-level couplings that may wait for the completion of an entire task, this model utilizes episodes or specific milestones within a session to update the robot's internal logic. For example, as the robot completes a sub-task, such as successfully navigating a single hallway or placing one component in an assembly line, it pauses briefly to integrate that specific experience before moving to the next segment. This structure increases the likelihood that early errors are corrected by the end of the session, rather than persisting until the robot is powered down and updated offline.

\subsection{Step Coupling}

The fourth position, step coupling, represents continuous learning throughout the interaction. The robot updates its internal knowledge representation incrementally at all times, resulting in a tight integration of learning and experience. In this structure, interaction and learning are maximally coupled. Behavior adaptation is potentially observable in real time. 

\subsection{Hybrid Coupling}

Hybrid coupling exists between every coupling level after singular coupling and above step coupling. Hybrid coupling is the mix of multiple learning processes within the same HRT design; it consists of at least one process at a given coupling level and another at that level or below. For example, a hybrid singular coupling HRT system may utilize step coupling for low-level motor control, allowing the robot to adjust its grip pressure in real-time, while maintaining singular coupling for high-level task logic, such as waiting until the end of the day to update its overall path-planning map based on the day’s obstacles.

Although the preceding definitions establish a conceptual taxonomy for coupling, the utility of this scale lies in its ability to categorize and evaluate diverse robotic systems relative to human learning theory. The following section outlines the practical application of this scale, providing a standardized methodology for researchers to benchmark robotic adaptability and identify the specific coupling thresholds required for complex human-robot collaboration.



\subsection{Application of Scale}

The application of this scale requires researchers to map the temporal frequency of learning updates against the progression of the HRT interaction. This process proceeds in two complementary steps: establish a ground truth of the robot learner's update behavior, and then reconstruct the teacher's perspective of that same process.

The first step is to establish the ground truth of the robot learner's knowledge updates. From the robot learner's perspective, this requires a systematic analysis of how often knowledge updates occur throughout the interaction and the conditions that trigger that update. Researchers should trace the HRT interaction to identify precisely when the robot's model changes state, whether that occurs after every timestep, at task-episode boundaries, or only upon session completion. The conditions causing these updates are equally important, as they reveal whether learning is driven by explicit teacher corrections, accumulated experience, environmental feedback, or some combination thereof. Together both dimensions allow the robot's actual coupling level to be objectively classified within the scale.

The second step shifts perspective from the robot to the human teacher. Because teachers generally do not have access to the robot's internal learning state, their understanding of when and whether the robot has learned is necessarily inferred from observable behavior. We therefore define the teacher's view of the learner's model as originating from the first explicit indication of a knowledge update in the robot's behavior. 

Through this we establish a baseline for the teacher's mental model of the robot learner to compare against the ground truth mental model of the robot learner. With this researchers can gather insights into different mechanics of the HRT interaction,
such as how the timing of learning, information exchange, and frequency of learning affect the learning interaction and the teacher's mental model. This perceptual threshold may not align with the ground truth update schedule. For example, a robot operating under step coupling may be updating continuously, yet a teacher may only recognize a learning event once the behavioral change becomes large enough to be observable.

By comparing the ground truth update schedule against the teacher's inferred mental model of the robot's learning, researchers gain a structured baseline for evaluating the dynamics of the HRT interaction. The gap between what the robot is actually learning and what the teacher believes the robot has learned, serves as a meaningful diagnostic signal. Discrepancies in this comparison can illuminate how the timing of learning, the exchange of information, and the frequency of knowledge updates collectively shape the teacher's mental model and, in turn, influence their instructional decisions. By making these dynamics explicit and measurable, the coupling scale provides researchers with a rigorous framework for understanding not only how robots learn from humans, but how humans adapt their teaching in response to the learning systems they perceive.

\section{Investigating the Effects of Coupling on Human-Robot Teaching}
We present our preliminary efforts on investigating the effects of coupling on HRT by applying the scale to a subset of the HRT literature. Namely, we applied the scale to the human-interactive robot learning literature and leveraged the reference list from \cite{baraka2026human}. This source was selected because its definition of HRT aligns with the focus of our work. This was not meant to be a comprehensive systematic literature review and instead focused on investigating the utility of applying the HRT coupling scale. 

A total of 195 papers were identified in the reference list. Papers were included in our analysis if they fit within the following inclusion criteria: 1) there was at least one robot interaction with at least one human, 2) a robot learning interaction occurs, 3) the human acts/communicates in ways that influence the robot's behavior. After applying the inclusion criteria, a total of 20 papers were retained.

These 20 papers were analyzed by placing each paper onto the scale from two perspectives. First, papers were positioned based on how the robot learning model was designed in the study, which served as the ground truth.
Second, papers were placed on the scale based on how coupling was perceived by the teacher, which consisted of identifying how the model updates of the learner were conveyed to the teacher. In addition to placing the papers on the scale, we collected data that was directly influenced by the human-robot teaching interaction, such as user trust, user workload, user preferences, interaction quality, and performance.


This data was then explored by identifying the distribution of papers across our scale grouped by their ground truth coupling and how coupling was perceived by the teacher. Through this categorization scheme, we then investigated patterns in outcomes within and across coupling positions, as well as misalignment in coupling between the ground truth and what the teacher perceived.



\section{Findings}


\subsection{No Coupling}

\subsubsection{Teacher's illusion of perceived learning}
No coupling systems are defined by the absence of a link between HRT interaction and robot learning. Despite this, no coupling HRT systems convincingly simulate coupled learning from the teacher's perspective. For example, in \cite{nagai2008toward}, \cite{schrage2024interactive}, and \cite{vollmer2014robots}, pre-scripted behaviors, gaze, and responsiveness create the appearance of adaptation. This reveals that teachers infer learning primarily from behavioral cues rather than actual model updates. Across \cite{nagai2008toward}, \cite{schrage2024interactive}, and \cite{vollmer2014robots}, behavioral cues are sufficient to drive teacher adaptation, leading teachers to modify demonstrations, exaggerate actions, and adjust timing. However, this produces a one-sided adaptation loop in which teacher effort accumulates without influencing future robot behavior, effectively turning interaction into an overhead cost rather than a mechanism for improvement.

\subsubsection{Teacher's illusion of influence}
At the same time, teachers often retain control over the timing and form of feedback \cite{luo2023rlif}, \cite{schrage2024interactive}, \cite{vollmer2014robots}. However, this control has no lasting consequence due to the absence of coupling. Robot behavior across these systems remains fixed or pre-scripted whether through ignored corrections, replayed trajectories, or static baselines. This results in consistent but non-improving responses that ultimately decouple teachers' perceptions of the robot's learning efficacy from its actual learning outcomes..

\subsubsection{Feedback modality richness masks functional stagnation}
Furthermore, the richness of feedback modalities such as physical correction, demonstration, mouse intervention, and verbal or tactile reinforcement, does not compensate for the absence of coupling. This disconnect produces systematic misalignment between teacher expectations and system capabilities, as seen in \cite{schrage2024interactive} and \cite{vollmer2014robots}, where users develop false mental models of learning based on perceived responsiveness. Ultimately, no coupling systems do more than just fail to support learning; they create an interface that actively masks a system’s functional stagnation.

\subsection{Singular Coupling}
\subsubsection{Singular coupling reduces data diversity and learning performance}
Singular coupling limits training data diversity and reduces learning performance. This is largely because the teacher lacks visibility into how the model updates its behavior in response to provided demonstrations, preventing the system from shaping or steering teacher input toward more informative demonstrations. In \cite{biyik2022learning}, inverse reinforcement learning (IRL) was used in a singular learning condition, which led to limited and less robust data. Later, the authors modified the IRL to use iterative coupling preference queries to learn human preferences. This resulted in a more performant model and better user satisfaction, while demonstrating that the addition of iterative coupling to the interaction can enhance the quality of data collected. Similarly, Hou et al.'s work on improving data collection \cite{hou2023shaping} compared two singular coupling conditions: active and natural. In the active condition, the robot repositioned between trials, creating the appearance of iterative learning, while in the natural condition the robot remained stationary. Participants in the active condition perceived greater diversity in the collected demonstrations and placed greater importance on producing diverse data. These differences were attributed to the distinct outcomes produced by the robot's repositioning behavior. In \cite{hou2024give}, the Deep Deterministic Policy Gradient (DDPG) singular condition generated lower-quality, redundant demonstrations compared to conditions where the agent iteratively learned and requested demonstrations. Finally, in \cite{chen2023fast}, the only singular condition without pretraining (i.e., teaching from scratch) performed worse, while policy mixture conditions with pre-training improved performance without altering the interaction. These findings suggest that singular coupling fails to structure feedback toward informative demonstrations, making data diversity a key bottleneck for learning. Demonstrations tend to be redundant, overly safe, or suboptimal, resulting in limited state-space coverage and poorer generalization.

\subsubsection{Mental model misalignment of robot knowledge state}
Singular coupling also disrupts the teacher’s ability to build an accurate mental model of the robot learner. In \cite{hou2023shaping}, participants in the active condition placed higher importance on data diversity when compared with the natural condition. This suggests that interaction structure influenced their priorities and understanding of the robot learner. In \cite{hou2024give}, the absence of displayed knowledge updates led to redundant demonstrations, whereas feedback conditions reduced mental workload and accelerated learning. Taken together, these findings suggest that interaction structure shapes how teachers infer the robot’s knowledge state. More importantly, when the interaction structure does not provide a mechanism for a teacher to infer a robot's knowledge state, it leads to inefficient and incorrect human teaching behavior.

\subsubsection{Temporal separation creates inefficiency}
Singular coupling enforces a strict “teach first, learn later” sequence, limiting opportunities for co-adaptation between teacher and learner. In \cite{hou2024give}, the DDPG singular condition, where all demonstrations were collected before learning, was slower than its hybrid counterparts and imposed higher mental workload on participants. This resulted in redundant inputs and longer completion times. In \cite{biyik2022learning}, supplementing singular demonstrations with iterative learning improved performance and highlighted the benefit of integrating learning with teaching rather than strictly separating them. Similarly, in \cite{chen2023fast}, the pretrained singular conditions that reduced reliance on the teaching phase mitigated inefficiencies, whereas the teaching from scratch condition revealed the negative impact of temporal separation on learning outcomes. By preventing real-time correction, clarification, and refinement, singular learning can lead to over-teaching and lower-quality data. These findings suggest that temporal separation constrains both performance and teaching efficiency, particularly in settings that rely heavily on teacher-generated demonstrations. 

\subsubsection{Breaking free from singularity results in learning gains}
Singular coupling is not inherently ineffective, but incomplete. It lacks mechanisms to structure teacher input and, in turn, limits the system’s ability to fully leverage human expertise. This is apparent in several studies where its augmentation consistently improved outcomes. For example, in \cite{biyik2022learning}, hybrid learning (i.e., singular demonstration plus iterative queries) strongly outperformed pure singular learning. In \cite{hou2024give}, step plus iterative approaches outperformed the singular conditions in accelerating learning and reducing teacher workload. In \cite{hou2023shaping}, the singular active condition increased data diversity and resulted in better performance. In \cite{chen2023fast}, aggregating multiple singular demonstrations through pretraining in the policy mixture conditions saw improved performance relative to learning from scratch. Even when the interaction appeared identical, as in \cite{chen2023fast}, adding additional sources of singular learning or pre-trained policies improved outcomes. 

This suggests that these additions to singular coupling such as queries, movement, feedback requests, and supplementary learning phases can dramatically enhance learning outcomes through more informative and diverse input during teaching. While interaction-based mechanisms can alleviate these limitations, comparable improvements may also be achieved by aggregating multiple demonstration sources as done in \cite{chen2023fast}. This suggests that the primary shortcoming of singular coupling is its inability to ensure diverse and informative training data within a fixed teaching phase, rather than a lack of iteration alone.

\subsection{Iterative Coupling}
Iterative coupling transforms HRT interaction from a series of discrete demonstrations into a co-adaptive process. In this paradigm, teaching is an ongoing regulation of behavior in which robot learning and teacher strategies recursively influence each other. Iterative coupling therefore shapes both robot performance and teacher adaptation, establishing a bidirectional learning dynamic.

The specific channel through which iteration is enacted whether through evaluative feedback, physical interventions, natural language, or continuous demonstrations dictates the type of interaction. For instance, systems based on repeated evaluation or preference queries tend to exhibit feedback with high consistency but limited expressiveness \cite{vollmer2018user,myers2023active}. In contrast, iterative linguistic feedback \cite{yang2024trajectory} provides high expressiveness and efficiency, while continuous demonstrations \cite{nagai2008toward} offer a more naturalistic flow that is frequently hampered by ambiguity.

\subsubsection{Cognitive effects of iterative coupling}

Crucially, iterative coupling induces a co-adaptive loop in which the teacher’s strategy adapts along with perceived robot learning. Teachers may begin to value data diversity \cite{hou2023shaping}, simplify demonstrations based on robot success \cite{vollmer2014robots}, or develop specific preferences for feedback modalities \cite{yang2024trajectory}. Novice participants showed emergent task understanding during iterative teaching, including discovery of structural task properties while teaching the robot \cite{adamson2021we}. We distinguish between system-regulated iterations such as when the teacher acts as an evaluator of system-generated actions through preference queries \cite{myers2023active}, and teacher-regulated iterations where the human assumes a more supervisory role by evaluating and correcting robot behavior through interaction-driven feedback \cite{losey2022physical}. This control structure directly impacts how teachers construct their understanding of the robot's intelligence. 

Importantly, the perception of iteration and learning may also influence teacher behavior, even when system-level learning is limited or not directly observable. This suggests that coupling should be understood not only as a property of the system design, but also a cognitive construct shaped by the human teacher's interpretation of interactions.

\subsection{Iterative Hybrid Coupling}
Iterative hybrid coupling occurs when multiple iterative or iterative and lower-level couplings coexist within a single robot learning system. For example, in \cite{biyik2022learning}, a teacher provides an initial demonstration and follows with a subsequent preference query. Similarly, in \cite{wang2021predicting}, teachers provide demonstrations and labeling before transitioning to online deployment learning.

\subsubsection{Hybrid systems shift the teacher’s role}
Hybrid coupling often consists of multiple distinct learning systems, each of which can alter the teacher's role within the interaction.  In \cite{biyik2022learning}, the teacher transitions from a demonstrator to an evaluator and in \cite{wang2021predicting} the teacher moves from a demonstration and labeling role to an interaction partner of the learner. Across both, hybrid coupling shifts teachers away from open-ended demonstrations toward increasingly structured forms of interaction.

\subsubsection{Hybrid coupling emerges to address limitations of singular learning}
A key driver of this structure is the limitation of singular coupling. Hybrid approaches emerge as a response to insufficient data diversity, where initial demonstrations fail to adequately cover the task space. In \cite{biyik2022learning}, this is explicit where singular coupling IRL suffers from limited training diversity and the hybrid approach improves robustness through structured queries. \cite{wang2021predicting} implicitly reflects the same pattern, where early demonstrations are partial and later online interaction expands coverage. Importantly, the performance gains observed in both works suggest that hybrid learning is not simply adding more data, but introducing structured guidance into the teaching process.

\subsubsection {Hybrid coupling can appear fragmented or unified in practice}
While hybrid coupling learning systems are often unified, it is not always perceived as such by teachers. In cases like \cite{biyik2022learning} and \cite{wang2021predicting}, the teacher experiences learning as distinct sequential activities (e.g., demonstrating then answering queries, teaching then interacting). This can be described as fragmented because hybrid learning is experienced as separate learning processes rather than a single integrated process.

However, this phase-based perception is not consistent across all hybrid systems. In \cite{liang2024learning}, hybrid coupling becomes perceptually collapsed. Additionally, the two distinct iterative processes of fast online adaptation and slow offline updating are experienced as a single continuous loop. This reveals a critical distinction: perceived coupling is not determined by the underlying learning algorithm, but by the visibility of learning complexity within the interaction. When learning processes are explicitly exposed, hybrid coupling is structured and multi-staged. When hidden, multiple learning mechanisms collapse into a single perceived mode despite remaining functionally distinct. This creates the appearance of a simpler learning system.

\subsubsection{Hybrid coupling introduces a trade-off between transparency and effort}
This variability introduces a fundamental design trade-off in hybrid systems. Visible hybrid coupling, as seen in \cite{biyik2022learning} and \cite{wang2021predicting}, promotes transparency, resulting in the teacher gaining a greater understanding of the structure of learning. However, this is often at the cost of increased interaction effort (e.g., answering queries, labeling data). In contrast, when learning mechanisms remain hidden, hybrid coupling preserves interaction simplicity and enables complex learning processes without increasing cognitive load on the teacher \cite{liang2024learning}. Thus, hybrid coupling operates along a spectrum. A more visible hybrid coupling is structured and interpretable, whereas a hidden hybrid coupling is lower effort but less transparent to the teacher. This highlights a core tension in human-robot teaching. Increasing learning complexity does not require increasing interaction complexity, but often does so at the expense of teacher awareness and interpretability of the learning process.

\subsection{Step Coupling}
Step coupling is defined by a learning system that operates in a continuous learning manner. Ideally, the robot updates its knowledge at every state of the interaction. However, within the reviewed literature, no system implemented true step coupling. The closest appears in \cite{vollmer2014robots}, where  robot gaze cues created the appearance of ongoing learning despite the absence of actual updates to the robot's internal model. This leads teachers to form dynamic mental models of the robot learner and adapt their demonstrations accordingly. While this perceived step coupling increased engagement and promoted adaptive teaching behavior, it also introduced the possibility of misalignment between teacher expectations and robot capabilities. As a result, teachers may overestimate the robot’s knowledge,  alter demonstrations and demonstration timing, or expend unnecessary effort based on perceived learning rather than actual learning. This reveals a critical tension and design consideration for step coupling: perceived step coupling can shape teaching behavior and facilitate more natural, collaborative interactions even when no underlying learning occurs. Consequently, feedback cues should be carefully aligned with the actual learning processes to avoid misleading teachers about the learning capabilities, unless distortion of the teacher's mental model is itself an intentional design objective. 

\subsection{Hybrid Step Coupling}
\subsubsection{Hybrid step coupling enables concurrent learning processes}
In contrast to preceding couplings, hybrid step coupling sits at the highest point of our scale. In \cite{hou2024give}, the robot continuously learns during task execution while incorporating teacher feedback during interaction. Similarly, in \cite{kessler2019active}, learning occurs throughout a stacking task while feedback is provided either through teacher-initiated input or system-requested queries. In \cite{christofi2024uncovering}, Q-learning learns continuously alongside policy-shaping. These works are therefore distinct from earlier hybrid formulations in that learning is not divided into distinct processes, but emerges from the real-time coupling of autonomous and interactive learning.

\subsubsection{Feedback timing becomes a defining constraint in hybrid step systems}
Contrasted with step coupling, in hybrid step coupling, learning is no longer purely continuous, but structured around key moments where feedback is more informative for learning. In \cite{hou2024give}, feedback is requested at specific states or episodes. In \cite{kessler2019active}, two conditions rely on algorithmically triggered feedback while one remains teacher-initiated. In \cite{christofi2024uncovering}, feedback is constrained through interaction budgets limiting the number of demonstrations and evaluative actions. 

Although these systems share the same underlying structure of concurrent learning and interaction, they differ in who controls the feedback and to what degree they control its timing. This reveals a consistent pattern: the tighter the alignment between feedback timing and the robot’s learning process, the lower the redundancy and the higher the efficiency. For instance, \cite{hou2024give} shows that purely singular learning leads to redundant demonstrations, whereas integrating continuous learning with real-time feedback reduces this effect. \cite{kessler2019active} and \cite{christofi2024uncovering} similarly constrain feedback, but through different mechanisms - structured interfaces in the former, and interaction budgets in the latter. These findings suggest that performance gains are driven less by the amount of human input and more by its placement within the learning process.

\subsubsection{Feedback integration, not modality, defines hybrid step coupling}
Importantly, hybrid step coupling is defined not by feedback modality, but by how feedback is integrated into an ongoing learning process. Across studies, feedback ranges from fine-grained demonstrations \cite{hou2024give} to binary evaluative signals \cite{kessler2019active}, \cite{christofi2024uncovering}, yet the underlying structure remains consistent: feedback functions as a regulatory signal within an ongoing learning process.  

This shifts the role of the teacher from the primary source of knowledge to a learning process regulator, providing targeted corrections, validations, or policy-shaping signals while the system maintains responsibility for continuous learning. As a result, hybrid step coupling reduces misalignment between teacher and system expectations and enables more efficient, lower-burden interaction. Together, these factors indicate that hybrid step coupling is primarily defined by the temporal regulation of feedback within continuous learning, rather than the form of feedback itself.

Overall, hybrid step coupling is characterized by concurrent learning processes, event-driven feedback, and system-mediated control of interaction timing. These properties support more efficient and scalable human-robot teaching dynamics.

\section{Conclusion}
We introduce a new taxonomy to review HRT papers based on the temporal and behavioral dynamics of the learning process. We introduce frequency of learning as a construct to categorize papers and define it by the density of coupled robot knowledge acquisitions and learning updates. We examine varying forms of coupling, ranging from no coupling to step coupling. We then analyze how these positions on the scale shape both robot learning outcomes and teacher behavior.

\bibliography{ref}

\end{document}